\documentclass[abstract]{nldl}

\renewcommand{\DOI}{12345}

\usepackage[utf8]{inputenc}
\usepackage{url}
\usepackage{graphicx}
\usepackage{authblk}
\usepackage{amsmath}
\usepackage{booktabs, subcaption}

\usepackage{algorithm}
\usepackage{algorithmic}
\usepackage{scalefnt}

\usepackage[square,numbers]{natbib}

\usepackage{soul} 

\usepackage{hyperref}

\title{Encoding Binary Events from Continuous Time Series in Rooted Trees using Contrastive Learning}

\author[1]{Tobias Engelhardt Rasmussen\thanks{Corresponding Author: tenra@dtu.dk}}
\author[1]{Siv Sørensen}
\affil[1]{Technical University of Denmark}

\date{\vspace{-5ex}}

\begin{document}
\nldlmaketitle

\begin{abstract}  
Broadband infrastructure owners do not always know how their customers are connected in the local networks, which are structured as rooted trees. A recent study is able to infer the topology of a local network using discrete time series data from the leaves of the tree (customers). In this study we propose a contrastive approach for learning a binary event encoder from continuous time series data. As a preliminary result, we show that our approach has some potential in learning a valuable encoder.
\end{abstract}
\section{Introduction}
Hybrid-Fiber Coaxial (HFC) networks are used to connect users to the internet using coaxial cables in which data is transmitted using radio frequency (RF) signals. Each HFC network is connected to the optical backbone grid through a local singular conversion node, called a CMC, where the signal is converted from coaxial to optical and vice versa. Within each HFC network each customer is connected to the CMC through a sequence of cable amplifiers and cable splitters arranged in a tree-like structure. Here, the customers represent the leaf nodes, the cable splitters the internal nodes, and the CMC the root node \cite{hfc}.

Due to digitization, complexity, and time, some degree of the cabling, hereinafter referred to as the \textit{topology}, of most HFC networks is unknown to the infrastructure owner. This is problematic since knowing the topology is crucial for network maintenance staff when localizing and resolving network faults. Incomplete topology records on average lead to a significant increase in both resolution and driving time for the maintenance staff.


A recent study, currently under review, shows that it is possible to infer a missing topology with high accuracy using discrete time series data collected only at leaf level \cite{siv}. The authors do this by considering both a \textit{Maximum Parsimony} optimality criterion \cite{fitch1971toward} as well as geographic distances between components of the network. The most \textit{parsimonious} tree topology is the topology which can explain the observed leaf data in terms of \textit{fewest} data mutations. I.e. given the RF signal is initially transmitted through a singular root node, one can reconstruct the most probable evolution of the signal throughout the network branches at each time step and count the number of changes \cite{fitch1971toward, parsimony}. This number is often referred to as the \textit{parsimony score} of a proposed tree topology.
However, this approach relies both on the observed data being discrete, as well as distinguishable data mutations, hereinafter \textit{events}, occurring on most branches of the network over time.

Our work focuses on extracting significant discrete events from continuous time series data, a task which, in this setting, is unsolved. For simplicity we initially consider binary events. Due to the parsimony score counting the number of mutations needed to explain the observed leaf data, learning an encoder using the parsimony score as a loss function is not feasible. No mutations are needed to explain identical data, so the model will simply encode the same data point for all leaf nodes.

In this work we propose using the parsimony score to train a binary event encoder using a contrastive approach \cite{contrastive}. As a preliminary result, we aim to analyse the feasibility of the proposed approach using simulated (perfect) data, where all possible topologies are known and unique events are simulated on all network edges. We will let this approach be the foundation for future work.


\section{Proposed Method}
We propose a contrastive approach building on a Siamese network\cite{siamese} where we will let the discrimination be determined by the parsimony score. Our approach consists of a Siamese network that encodes the continuous time series of all the modems into a sequences of bits. The parsimony score is calculated across multiple topologies and used to minimize the score of the true topology while simultaneously maximising the difference in parsimony scores using different topologies. Our approach is illustrated in \autoref{app:contrastive_approach}. For the encoder model we propose using a causal convolutional deep neural network inspired by Franceschi \textit{et. al} in \cite{causal_conv}. However, we do not pool the time dimension (the desired output is a time series), but instead send each time point through a sigmoid-function.

Let $\mathcal{P}(\mathbf{X}, G)$ be the vector of parsimony scores for each time point/step of the encoded bits $\mathbf{X}$ using the graph $G$. We initially propose the loss function:
\begin{equation}
\scalefont{.95}
\begin{split}
    &L(\mathbf{X}, G^p, G_{1}^n, \dots G_{K}^n) = \sum \mathcal{P}(\mathbf{X}, G^p) \ + \\ &\frac{\alpha}{  2}\sqrt{\sum_{g_1\in \mathcal{G}} \sum_{g_2\in \mathcal{G}\backslash g_1} \frac{1}{\left(\| \mathcal{P}(\mathbf{X}, g_1) - \mathcal{P}(\mathbf{X}, g_2)\|_2^2 + 1\right)^2}}
\end{split}
\end{equation}
where $G^p$ is the true topology, $G^n_\bullet$ are the $K$ negative (sampled) topologies, and $\mathcal{G} = \{G^p, G^n_1, \dots, G^n_K\}$. This loss function considers the difference between resulting parsimony scores using all pairwise topology samples, rewarding large score differences and punishing small ones.
\section{Evaluation}
In order to validate the feasibility of the proposed algorithm we run it on simulated data and present the results.

For a tree with four internal nodes whereof one is a root node, there exist 16 different permutations of the topology. All possible permutations of the tree are provided in the \autoref{app:all_topos}. We simulate both the network topology and number of modems for each splitter along with the continuous time series leaf data. We use the AR(2) model\cite{time_series_analysis} as a base signal inspired by \cite{ar2}. We create perfectly distinguishable events by splitting the time points into chunks (with a start and end time) and for each chunk we sample an internal edge and modify the signal of the affected modems, below this edge in the network, accordingly. The sampling procedure is described in detail in the \autoref{app:simulation}.

We simulate 1,000 observations whereof 500 will be used as a separate test set. For each observation we simulate time series of length 4,000 that we split into 40 chunks for simulating events. We train the model experimentally setting the weight-parameter to $\alpha=2.0$. \autoref{tab:topo_results} shows the fraction of observations for which the true topology leads to the lowest parsimony score, respectively, the second and third lowest parsimony score of all possible topologies.
\begin{table}
\caption{Results of our encoded model}
\begin{subtable}[t]{.5\linewidth}
    \centering
    \caption{Fraction of observations in the test set for which the true topology leads to the lowest parsimony score.}
    \label{tab:topo_results}
    \vspace{.1cm}
    \begin{tabular}{c c c} 
       Top-1  & Top-2 & Top-3 \\ \toprule
        0.748 & 0.956 & 0.998 \\ \bottomrule
    \end{tabular}
\end{subtable}
\hspace{\fill}
\begin{subtable}[t]{.46\linewidth}
    \centering
    \caption{Event accuracy reported as the mean and standard deviation across the 500 observations.}
    \label{tab:event_results}
    \vspace{.1cm}
    \begin{tabular}{c c}
        Mean & Std. \\ \toprule
        0.600 & 0.0967 \\ \bottomrule
    \end{tabular}
\end{subtable}
\end{table}
In order to check how well the encoder learns the encoded events, we also report the accuracy of the event encoder in \autoref{tab:event_results}.
\section{Discussion and Conclusion}
Our results show that for around 75\% of the test cases, the encoded time series resulted in the true topology leading to the minimum parsimony. In almost every case the encodings resulted in the second or third lowest parsimony. This means that our approach has some potential in being able to learn valuable binary encodings for the topology reconstruction problem.

The event accuracy is fairly low and we hypothesize that this is partly due to a delay in the event detection. We use a causal convolutional encoder which means that the encoder only has access to information prior to a given time. Due to the urgency of the topology reconstruction problem not being imminent, we believe that an ordinary convolutional encoder could lead to better results in future work.

\clearpage
\bibliographystyle{abbrvnat}
\bibliography{references}

\appendix
\section{Illustration of porposed approach} \label{app:contrastive_approach}
\begin{figure*}
    \centering
    \includegraphics[width=\linewidth]{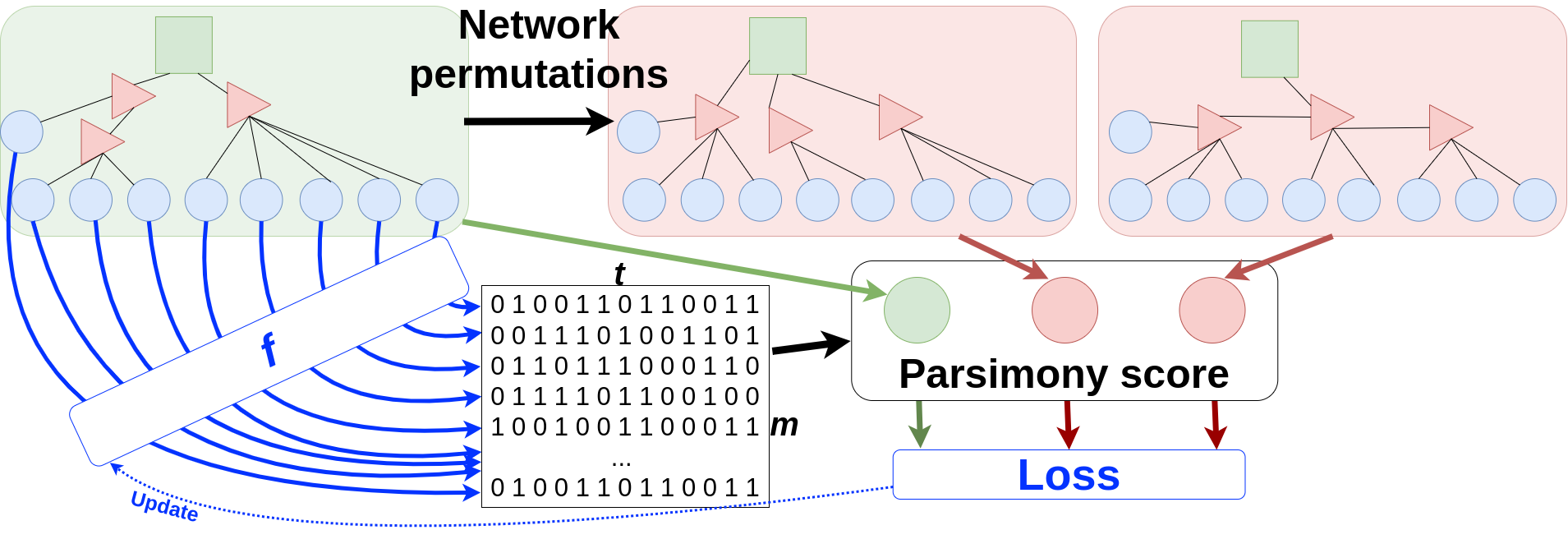}
    \caption{Illustration of the proposed algorithm. A Siamese network is trained using the parsimony score on a positive (true topology) sample (green) and a set of negative samples (red). The time series from each of the $m$ customers are sent through the same encoder $f$ that encodes continuous time series into an $m\times t$ matrix of binary events. Using this matrix and each of the different topologies, the parsimony scores can be computed. Based on these the loss is calculated and used to update the encoder.}
    \label{fig:contrastive_approach}
\end{figure*}

Our proposed approach is illustrated in \autoref{fig:contrastive_approach}.

\section{Possible tree topologies using four internal nodes} \label{app:all_topos}
\begin{figure*}
    \centering
    \includegraphics[width=.5\linewidth]{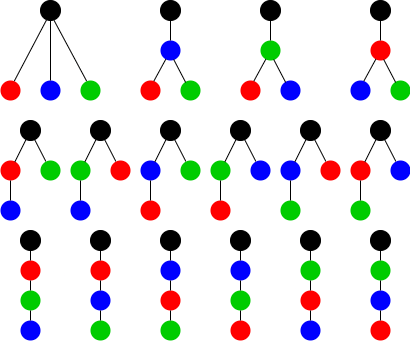}
    \caption{All possible topologies for a network with 4 internal nodes. The black node is the root and the colored dots are the splitters. Notice that each of the splitters also have leaf-nodes as children that are not shown for simplicity.}
    \label{fig:all_topos}
\end{figure*}

All possible topologies for a tree with four internal nodes is illustrated in \autoref{fig:all_topos}. In this illustration we do not show the leaf nodes for simplicity.

\section{Data simulation procedure} \label{app:simulation}
We sample each observation using the following approach.
\begin{itemize}
    \item Sampling a random topology from the 16 candidates shown in \autoref{app:all_topos}
    \item Sample a number of modems $M\in [50, 100]$ and distribute these to the three internal nodes (not the root)
    \item Set $\mu_{m,t}=0$ for all modems
    \item Divide time stamps into a set of chunks of a predertermined size $\tau$ ($t_i, t_{i+\tau}$)
    \item For each chunk sample an internal edge on which an event occurs. Determine the affected modems ($M^*$) and timepoints ($T^*$) and set: 
    $$\mu_{m, t}=3, \quad m\in M^*, t\in T^*$$
    \item Sample a signal using the AR(2) model:
    $$x_t = 0.6x_{t-1} -0.5x_{t-2} + \epsilon_t, \quad x_{0}=x_1=0$$
    where $\epsilon_t \sim \mathcal{N}(\mu_t, 0.5^2)$
\end{itemize}

\end{document}